\documentclass[letterpaper, 10 pt, conference]{ieeeconf}  

\IEEEoverridecommandlockouts                              
\usepackage{graphics} 
\usepackage{epsfig} 
\usepackage{amsmath} 
\usepackage{amssymb}  
\usepackage{epstopdf}
\usepackage{cite}

\title{\LARGE \bf
A Nonparametric Adaptive Nonlinear Statistical Filter
}

\author{Michael Busch and Jeff Moehlis
\thanks{M. Busch is a graduate student of Mechanical Engineering,
        University of California, Santa Barbara, USA
        {\tt\small mbusch@engr.ucsb.edu}}%
\thanks{J. Moehlis is with the Department of Mechanical Engineering, University of California, Santa Barbara, USA
        {\tt\small moehlis@engr.ucsb.edu}}%
}

\begin{document}

\maketitle
\thispagestyle{empty}
\pagestyle{empty}

\begin{abstract}

We use statistical learning methods to construct an adaptive state estimator for nonlinear stochastic systems. Optimal state estimation, in the form of a Kalman filter, requires knowledge of the system's process and measurement uncertainty. We propose that these uncertainties can be estimated from (conditioned on) past observed data, and without making any assumptions of the system's prior distribution. The system's prior distribution at each time step is constructed from an ensemble of least-squares estimates on sub-sampled sets of the data via jackknife sampling. As new data is acquired, the state estimates, process uncertainty, and measurement uncertainty are updated accordingly, as described in this manuscript.

\end{abstract}

\section{Introduction}

Let us consider a continuous nonlinear model that contains model uncertainty in the form of a stochastic forcing term, and  is measured at discrete instances of time $t_k$:
\begin{subequations} \label{eqn:SDEmodel}
\begin{align} 
dx &= f(t,x)dt + \sqrt{Q}dw, \label{eqn:SDEa} \\
y(t_k) &= h(x(t_k)) + \sqrt{R}N(0,1), \label{eqn:SDEb}
\end{align}
\end{subequations}
where $x \in \mathbb{R}^n$ represents the state of the system, $f(t,x):\mathbb{R} \times \mathbb{R}^n \rightarrow \mathbb{R}^n$ is the deterministic evolution of the states, $h(x):\mathbb{R}^n \rightarrow \mathbb{R}^m$ is a function that maps $x$ to the discrete-time measured output $y\in \mathbb{R}^m$, $dw$ describes a vector Wiener process with mean zero and unit variance, and $N(0,1)$ represents a normally distributed random variable with zero mean and unit variance. It is also noted that the covariance matrices $Q$ and $R$ are positive semi-definite symmetric matrices, and their square roots exist and can be computed using a singular value decomposition~\cite{Strang2011}.

Since its discovery, the Kalman filter, in both its linear and nonlinear forms, has been an effective model-based noise filter that relies on an assumed known deterministic model with additive noise~\cite{Simon2006}.  However, when either parameters of the model or noise variances are unknown, which is common in tasks where model identification and state estimation must occur simultaneously, the Kalman filter is likely to diverge~\cite{Schlee1967,Fitzgerald1971}.

To prevent divergence, various tuning procedures exist for finding the best estimates of process and measurement noise for given a Kalman filter~\cite{Aakesson2007,Oshman2000,Powell2002,Ding2012,Lau2011}. Kalman filter tuning typically involves minimizing the measurement error over iterations of the Kalman filter, with the process and measurement covariances as the free variables. For nonlinear systems, this type of tuning procedure requires that at each optimization iteration, the gradient of a complete time sequence of Kalman filter iterations is taken with respect to all of the free variables. Hence, these methods are computationally costly, and are susceptible to converging to suboptimal local minima of their objective functions.

Adaptive algorithms have also been developed to allow the Kalman filter to converge on the correct noise values in an online manner~\cite{Song2008,Fathabadi2009,Forbes2012,Hajiyev2013, Jiang2007,Karasalo2011,Kosanam2004,Li2012,Mehra1972,Sarkka2013, Sarkka2009}. Much effort has been given to developing adaptive methods for nonlinear systems because online computation is in the spirit of the Kalman filter. Adaptive methods for linear systems have seen much success over the years~\cite{Mehra1972}, but their formulation is limited to the linear case and does not extend to nonlinear systems in general. For nonlinear systems, adaptive strategies have been implemented for the main variants of the nonlinear Kalman filter: the extended Kalman filter (EKF)~\cite{Karasalo2011}, and the unscented Kalman filter (UKF) ~\cite{Song2008,Hajiyev2013,Jiang2007}. However, for the adaptive EKF and UKF methods, convergence performance has yet to be rigorously generalized, and is sensitive to the initial estimates of the unknown parameters.

To overcome these challenges associated with implementing adaptive nonlinear Kalman filters, we propose that the unknown state and parameter distributions of the given model can be estimated by an ensemble of least-squares regression (LSQ) estimates on the known data. Jackknife sampling methods~\cite{Miller1974,Efron1981,Shao1989a} can be used to generate the ensemble of LSQ estimates~\cite{Shao1992}, and this ensemble generation procedure can then be made adaptive (in a Markov-Chain sense) by taking advantage of how jackknife sampling assimilates newly acquired data into the model. The formula for a statistical Kalman filter can then be used to infer the unknown process uncertainty and measurement noise covariance matrices from ensemble estimates at each step. After the unknown quantities of the stochastic model have converged, the adaptive procedure can be stopped, and a standard nonlinear Kalman filter can be implemented to take over the state estimation process. 

Although our approach is supported by the theory behind ensemble Kalman filtering (EnKF)~\cite{Evensen1994,Evensen2003}, our adaptive method of assimilating the data is original, as well as our application of jackknife sampling to generate ensemble members. Particle filters and the EnKF both make assumptions on the sampling distribution of states, and typically rely on Markov-Chain Monte Carlo (MCMC) simulation to generate ensemble members and deduce ensemble statistics of the states. We show that by using LSQ estimation in conjunction with jackknife sampling of the known data, a sampling distribution and ensemble statistics can be acquired without making any assumptions of the sampling distribution nor having to run a high number of MCMC simulations. Furthermore, we describe how our adaptive method can be implemented in a parallel setting, and with a fixed number of computations at each update step.


Therefore, the aim of this manuscript is to implement the techniques of jackknife variance estimators as they apply to least squares estimators, to construct an adaptive, nonparametric, and computationally efficient statistical nonlinear filter. To motivate our jackknife sampling LSQ approach for generating ensemble statistics, we shall present an overview of the derivation of a statistical Kalman filter in Section~\ref{sec:statistical_kalman}. In Section~\ref{sec:jackknife} we present a description of jackknife sampling methods, an adaptive jackknife sampling approach to assimilating new data, and how jackknife sampling can be used with LSQ estimation. We combine the results of Sections~\ref{sec:statistical_kalman} and~\ref{sec:jackknife} to construct a procedure in Section~\ref{sec:QandR} for estimating the process and measurement noise of the model~\eqref{eqn:SDEmodel}. We present an example application in Section~\ref{sec:example} to demonstrate the efficacy of our adaptive jackknife filter, and we summarize our conclusions and directions for future work in Section \ref{sec:conclusion}.


\section{Ensemble Kalman Filtering} \label{sec:statistical_kalman}

The ultimate objective of this manuscript is to develop a procedure to optimally estimate the states of a noisy nonlinear state-space model, when only a model structure and some observed measurements are provided. Since only a model structure is assumed, we shall attempt to estimate model parameters while simultaneously estimating model states. Optimal state estimation is often performed using a Kalman filter, which has many linear and nonlinear variants. For reasons that will be discussed later, we shall focus our attention on the EnKF and its formulation~\cite{Simon2006,Evensen1994,Evensen2003}. Here, we will describe the EnKF in order to motivate our approach for estimating the unknown model parameters in the next section.

\subsection{Model Uncertainty Propagation}

Let us consider a general nonlinear model that contains model uncertainty in the form of a stochastic forcing term
\begin{align} \label{eqn:SDE}
dx &= f(t,x)dt + g(x)dq,
\end{align}
where $x$ represents the state of the system, $f(t,x)$ gives the deterministic evolution of the states, $g(x)$ is a function that may depend on the states, and $dq=\sqrt{Q}dw$ describes a vector Wiener process with mean zero and covariance matrix $Q\delta (t)$. It is noted as a technical detail that since $g(x)$ is not an explicit function of $dq$, the Ito interpretation is used~\cite{Jazwinski2007}, and $\int^{t_k}_{t_{k-1}}{dw}=\sqrt{t_k-t_{k-1}}N(0,1)$. Thus, one can integrate \eqref{eqn:SDE} from $t_{k-1}$ to $t_k$ to obtain the distribution of $x(t_k)$ when the distribution $x(t_{k-1})$ is known. The probability distribution of $x(t_k)$ for a given initial point $x(t_{k-1})$ is
\begin{equation} \label{eqn:SDE_soln}
x(t_k) = F(t_{k},x(t_{k-1})) + g(x(t_{k-1}))\sqrt{Q \Delta t_k} N(0,1),
\end{equation}
where $\Delta t_k = (t_k - t_{k-1})$, and $F(t_{k},x(t_{k-1}))$ is the evolution operator that deterministically maps $x$ from time $t_{k-1}$ to $t_k$ according to the $dx=f(t,x)dt$ part of~\eqref{eqn:SDE}.

When $g(x)dq$ is normally distributed and forms a Markov process, it is shown in~\cite{Evensen1994} that it is possible to derive the Fokker-Planck equation to describe the time evolution of the probability density function $p(x,t)$ of the model state:
\begin{equation} \label{eqn:fokker-planck}
\frac{\partial p(x,t)}{\partial t} + \sum_i{\frac{\partial(f_i(t,x)p(t,x))}{\partial x_i}} = \frac{1}{2} \sum_{i,j}{\frac{\partial^2 (gQg^T)_{ij}}{\partial x_i \partial x_j}},
\end{equation}
where $f_i(t,x)$ is the $i^{th}$ component of $f(t,x)$, and $gQg^T$ is the covariance matrix for the model errors at time $t$.

The EnKF, as discussed in~\cite{Evensen1994} and~\cite{Evensen2003}, applies a Markov Chain Monte Carlo Method (MCMC) to solve~\eqref{eqn:fokker-planck}. The probability density $p(x,t)$ is represented by an ensemble of $N$ model states $x^{(i)}$ for $i \in \{1,\ldots,N\}$, and the ensemble prediction, by integrating model states forward according to~\eqref{eqn:SDE_soln}, is equivalent to using a MCMC method to solve~\eqref{eqn:fokker-planck}. Hence, there is no need to find an explicit form for the solution $p(x,t)$ of~\eqref{eqn:fokker-planck} because $p(x,t)$ can be sufficiently described by its ensemble statistics.

Since we assume no prior knowledge of the function $g(x)$, we shall simplify matters and take $g(x)=I_{n \times n}$ so that all of the model uncertainty is spatially invariant and entirely attributed to the process noise. Furthermore, we shall assume discrete measurements $y_k$ at times $t_k$, which have their own uncertainty that we shall assume to be normally distributed. For the remainder of the manuscript we shall assume the continuous-discrete stochastic model defined by~\eqref{eqn:SDEmodel}.

\subsection{A General Statistical Kalman Filter}

To help explain how the Kalman filter is implemented from an ensemble of nonlinear system realizations, we first introduce the Kalman filter. The following description for a general Kalman filter closely follows~\cite{Simon2006}, and is consistent with the EnKF of~\cite{Evensen2003}. However,~\cite{Evensen2003} assumes a linear mapping from the states to the outputs, but here we want to allow for nonlinear mappings, too.

We define the following variables at the discrete time instance $t_k$ of the latest measurement $y(t_k)$:
\begin{itemize}
\item $x(t_k)$ = true state value,
\item $\widehat{x}^-(t_k)$ = state estimate prior to measurement,
\item $\widehat{x}^+(t_k)$ = posterior state estimate,
\item $P^-(t_k) = E\left[(x(t_k)-\widehat{x}^-(t_k))([\cdots])^T\right]$,
\item $P^+(t_k) = E\left[(x(t_k)-\widehat{x}^+(t_k))([\cdots])^T\right]$,
\end{itemize}
where the $[\cdots]$ is shorthand notation for the term immediately to the left of it, so that the covariance matrices are written as $E\left[(z)([\cdots])^T\right] = E\left[(z)(z)^T\right]$. For notational convenience, we shall momentarily omit any explicit dependence on $t_k$ because all of the variables are understood to be implicitly evaluated at the same time instance $t_k$.

As described in~\cite{Simon2006}, the Kalman filter is defined as 
\begin{align}
\overline{\widehat{x}^+} &= \overline{\widehat{x}^-} + K\left(y-\overline{y}\right), \label{eqn:(10.97a)} \\
P_x^+ &= P_x^- - K P_{xy}^T, \label{eqn:(10.97b)} \\
K &= P_{xy}P_y^{-1}. \label{eqn:(10.96a)}
\end{align}
Here, the notation $P_{ab}$ denotes the cross-covariance of random variables $a$ and $b$. This choice of $K$ in \eqref{eqn:(10.96a)} minimizes the variance of the state estimates in \eqref{eqn:(10.97b)}, and \eqref{eqn:(10.97a)} is an unbiased estimator of the model states (i.e., $\overline{\widehat{x}^+}=\overline{x}$). We remark that \eqref{eqn:(10.97a)} has the Markov Property, and is only true when the Markov Property is true for each of its elements. In the next section we will discuss how one can use the ensemble output statistics to appropriately estimate $\overline{y}$, and prevent measurement bias from affecting the state estimate in~\eqref{eqn:(10.97a)}.

\subsection{Ensemble estimation of $P_x$ and $P_y$}

When integrating an ensemble of points forward in time according to~\eqref{eqn:SDE_soln}, the state covariance matrix $P_x^-$ depends on the distribution of those deterministic points and the stochastic forcing term. For notational convenience, let us denote $\widehat{x}^- = F(t_k,x(t_{k-1}))$. Since the ensemble mean is unbiased so that $\overline{\widehat{x}^-} = \overline{x}$, then an approximation for the prior ensemble covariance $P_x^-(t_k)$ becomes
\begin{align}
P_x^- &= E\left[ (x-\overline{\widehat{x}^-}) (\cdots)^T \right] \nonumber \\
&= E\left[ (\widehat{x}^- -\overline{\widehat{x}^-} + \sqrt{Q \Delta t_k} N(0,1)) (\cdots)^T \right] \nonumber \\
&= \frac{1}{N-1}\sum_{i=1}^{N}{\left[ (\widehat{x}_{(i)}^- -\overline{\widehat{x}^-} ) (\cdots)^T \right]} + Q \Delta t_k \nonumber \\
&= \widehat{P_x}^- +Q \Delta t_k, \label{eqn:Px_ensemble}
\end{align}
where $\widehat{P_x}^-$ is the sample ensemble covariance of the state prior distribution, and
\begin{equation}
\overline{\widehat{x}^-} = \frac{1}{N} \sum_{i=1}^{N}{\widehat{x}_{(i)}^-} \nonumber
\end{equation}
for the collection of $N$ ensemble members $\widehat{x}_{(i)}^-$.

To find the measurement covariance $P_y$ and cross-covariance $P_{xy}$, the process noise can be made an explicit term by taking a series expansion of $h(x(t_k))$ about $\widehat{x}^-$ at time $t_k$:
\begin{align}
h(x(t_k)) &= h(\widehat{x}^-+\sqrt{Q\Delta t_k}N(0,1)) \\
&= h(\widehat{x}^-) + Dh_{\widehat{x}^-} \sqrt{Q\Delta t_k}N(0,1) \nonumber \\ 
&+ \sum_{n=2}^\infty{\frac{1}{n!}D^nh_{\widehat{x}^-}(\sqrt{Q\Delta t_k}N(0,1))^n}, \label{eq:4.3}
\end{align}
where $D^nh_{x^{(i)}}$ represents the $n^{th}$ vector derivative of $h$ about the point $x_{(i)}(t_k)$. From this we obtain
\begin{align}
P_y &= E\left[ (y-\overline{y}) (\cdots)^T \right] \nonumber \\
&= E\left[ (h(x(t_k))-\overline{h(x(t_k))} + \sqrt{R} N(0,1)) (\cdots)^T \right] \nonumber \\
&= E\left[ (h(\widehat{x}^-) + Dh_{\widehat{x}^-} \sqrt{Q\Delta t_k}N(0,1)-\overline{h(\widehat{x}^-)}\right. \nonumber \\
&+ \left.\sqrt{R} N(0,1)) (\cdots)^T \right] \nonumber \\
&= \frac{1}{N-1}\sum_{i=1}^{N}{\left[ (h(\widehat{x}_{(i)}^-) -\overline{h(\widehat{x}_{(i)}^-)}) (\cdots)^T \right]} \nonumber \\
&+ \frac{1}{N}\sum_{i=1}^{N}{Dh_{\widehat{x}_{(i)}^-}Q\Delta t_k Dh_{\widehat{x}_{(i)}^-}^T} + R \nonumber \\
&= \widehat{P}_y + \widehat{Q}_y + R, \label{eqn:Py_ensemble}
\end{align}
where $\widehat{P}_y$ is the sample ensemble covariance of the measurements and $\widehat{Q}_y$ comes from the stochastic forcing term. Similarly, one finds the cross covariance to be 
\begin{align}
P_{xy} &= E\left[(x-\overline{x})(y-\overline{y})^T\right] \nonumber \\
&= E\left[(\widehat{x}^- -\overline{x}) (h(\widehat{x}^-)-\overline{h(\widehat{x}^-)})^T\right] \nonumber \\
&= \frac{1}{N-1}\sum_{i=1}^{N}{\left[ (\widehat{x}_{(i)}^- -\overline{\widehat{x}_{(i)}^-}) (h(\widehat{x}_{(i)}^-) -\overline{h(\widehat{x}_{(i)}^-)})^T \right]} \nonumber \\
&= \widehat{P}_{xy}, \label{eqn:Pxy_ensemble}
\end{align}
where $\widehat{P}_{xy}$ is the sample ensemble cross-covariance and all additive noise terms vanish because they are mutually uncorrelated.

By substituting equations \eqref{eqn:Px_ensemble}, \eqref{eqn:Py_ensemble}, and \eqref{eqn:Pxy_ensemble} into equations~\eqref{eqn:(10.96a)} and \eqref{eqn:(10.97b)}, one obtains
\begin{align}
K &= \widehat{P}_{xy}(\widehat{P}_y + \widehat{Q}_y + R)^{-1} \\
\widehat{x}^+ &= \widehat{x}^- + K\left(y-\overline{y}\right) \label{eqn:ensemble_mean} \\
P_x^+ &= \widehat{P}_x^- +Q \Delta t_k -K(\widehat{P}_y + \widehat{Q}_y + R)K^T . \label{eqn:ensemble_var}
\end{align}

However, to implement this nonlinear statistical filter, we need to have quantities for $Q$ and $R$, which we propose can be estimated directly from the data, and without making any assumptions on their sampling distribution. We did make assumptions that the process and measurement noise terms are Gaussian, which we will find in the next section, is actually consistent with a least squares parameter estimation strategy.

\section{Ensemble generation and adaptive update} \label{sec:jackknife}

In order to implement a statistical Kalman filter, we need to obtain ensemble estimates of the state and output distributions. To do this, we can take a statistical sample of those distributions via \textit{jackknife sampling}~\cite{Miller1974,Efron1981,Shao1989a}, which has been shown to be a robust and computationally efficient way of estimating the sample distribution of a given population. By mapping the data to the state-space via LSQ estimation, we shall jackknife sample the known data in order to obtain the underlying sample distribution of the states and model parameters. The points that define the sample distribution of the states and model parameters are then treated as ensemble members for the statistical Kalman filter. We shall first explain jackknife sampling, its consistency properties and an adaptive update rule, and then apply jackknife sampling to LSQ estimation. 

\subsection{Jackknife Sampling}

Suppose we are given a sequence of $n$ data measurements $D_n=\{Y_1,\ldots,Y_n\}$, where $Y_i=(y_i,t_i)$ is defined for an observed response vector $y_i$ from a known input sequence of $t_i$ values. For the moment, let us fix the number of available data points $n$ and choose some fixed positive integer $d$. We shall describe the \textit{delete-d jackknife} estimator~\cite{Efron1981,Shao1989a}, which estimates the sample distribution of parameters by aggregating the least squares estimates on randomly chosen subsets of $r=n-d$ data points. Let $S_r$ be the collection of subsets of $\{1,\ldots,n\}$ that have size $r$. For $s=\{i_1,\ldots,i_r\} \in S_r$, let $\widehat{\theta}_s = \widehat{\theta}\left(Y_{i_1},\ldots,Y_{i_r}\right)$. The delete-$d$ jackknife estimator of $\text{var}(\theta_n)$ is defined as
\begin{equation} \label{eqn:jackknife-1a}
v_n=\frac{r}{dN}\sum_{s\in S_r}{\left(\widehat{\theta}_s-\theta_n\right) \left(\widehat{\theta}_s-\theta_n\right)^T},
\end{equation}
where $N=\binom{n}{d}$, and $\theta_n$ is the parameter estimate that explains all of the available $n$ data points. For a finite set of measurements, we can approximate $\theta_n$ by the arithmetic average of subsample means, which we call the jackknife estimate $\widehat{\theta}_n$, and define
\begin{equation} \label{eqn:jackknife-1b}
\widetilde{v}_n=\frac{r}{dN}\sum_{s \in S_r} {\left( \widehat{\theta}_s - \widehat{\theta}_n \right) \left( \widehat{\theta}_s - \widehat{\theta}_n \right)^T}
\end{equation}
with
\begin{equation*}
\widehat{\theta}_n = \frac{1}{N}\sum_{s \in S_r}{\widehat{\theta}_s}.
\end{equation*}

When $N$ is very large, the number of computations can be reduced by implementing techniques from survey sampling. For instance, take a simple random sample (without replacement) of size $m$ from $S_r$ (i.e., $S_m \subset S_r$). Compute $\widehat{\theta}_s$ for $s \in S_m$, and use
\begin{equation} \label{eqn:jackknife-2a}
v_n^s=\frac{r}{dm}\sum_{s \in S_m} {\left(\widehat{\theta}_s-\theta_n\right) \left(\widehat{\theta}_s-\theta_n\right)^T}
\end{equation}
and
\begin{equation} \label{eqn:jackknife-2b}
\widetilde{v}_n^s=\frac{r}{dm}\sum_{s \in S_m} {\left(\widehat{\theta}_s- \widehat{\theta}_n\right) \left(\widehat{\theta}_s- \widehat{\theta}_n\right)^T}
\end{equation}
with
\begin{equation*}
\widehat{\theta}_n = \frac{1}{m}\sum_{s \in S_m}{\widehat{\theta}_s}.
\end{equation*}
to approximate $v_n$ and $\widetilde{v}_n$, respectively. These approximations are called the \textit{jackknife-sampling variance estimators} (JSVE's)~\cite{Efron1981,Shao1989a}, and $m$ is the second-stage sample size. It is also noted that the pre-factor terms $r/(dN)$ and $r/(dm)$ are explained in~\cite{Efron1981,Shao1989a}, and mitigate the bias associated with estimating the variance from a finite sample.

In~\cite{Shao1989b}, it was shown that 
\begin{itemize}
\item (\cite{Shao1989b} Theorem 1) $\text{var}(v_n) = o\left(n^{-2}\right)$,
\item (\cite{Shao1989b} Theorem 2) $0 \leq \text{var}(v_n^s) - \text{var}(v_n) = O(m^{-1} \tau_n)$, for $\tau_n=E\left[\left(\theta_n-\theta\right)^4\right]$.
\end{itemize}
We remark that $\text{var}(v^s_n)$, $\text{var}(v_n)$, and $E\left[\left(\theta_n-\theta\right)^4\right]$ are well defined for jointly distributed random variables~\cite{Ghazal2000}, and are only needed here to prove asymptotic consistency of jackknife sampled distributions.


The authors of~\cite{Shao1989b} also show that choosing $m=n^\delta$ for some $\delta \geq 1$ is sufficient and has the same number of computations as the delete-1 jackknife estimator. If $m$ is much smaller than $N$, sampling with replacement for the second-stage sample will produce almost the same estimator as sampling without replacement, which further simplifies the sampling procedure and is nearly identical to bootstrap sampling. It is also important to note that these results do not necessarily rely on $m^{-1}\sum_{s\in S_m}{\theta_n} \rightarrow \theta$ as $m \rightarrow \infty$, which is a convergence result that we will further discuss next.

\subsection{Adaptive Jackknife Variance Estimator}


Although the estimates are conditioned on past data, we see that the ensemble jackknife estimates abide by the Markov property in the sense that they only rely on the previous ensemble measurement and the current ensemble measurement. When tracking only the mean and variance of the distribution, all of the previous ensemble members may be forgotten, as their statistics are sufficiently captured by the mean and variance.

Suppose another measurement is collected so that there are now a total of $n+1$ data points, and for computational reasons we want the values of $r$ and $m$ to remain the same as before. When constructing the basic form of our adaptive equations, it is important to define the mean and variance of the linear combination of two uncorrelated random variables $X_1$ and $X_2$. For $\mu_1=E[X_1]$, $\mu_2=E[X_2]$, $v_1 = \text{var}(X_1)$, $v_2 = \text{var}(X_2)$, and two constants $ a_1 , a_2 \in \mathbf{R}$ such that
\begin{equation}
X_3 = a_1 X_1 + a_2 X_2, \nonumber
\end{equation}
then
\begin{align}
E[X_3] &= a_1 \mu_1 + a_2 \mu_2, \label{eqn:exp_combo} \\
\text{var}(X_3) &= a_1^2 v_1 + a_2^2 v_2. \label{eqn:var_combo}
\end{align}
In this context, each jackknife estimate $\widehat{\theta}_{n}$ can be view as a combination of jackknife estimates $\widehat{\theta}_{n \in s}$ that include the $n^{th}$ data point, and those that do not $\widehat{\theta}_{n \not\in s}$:
\begin{equation} \label{eqn:combination}
\widehat{\theta}_{n} = a_1 \widehat{\theta}_{n\in s} + a_2 \widehat{\theta}_{n \not\in s},
\end{equation} 
where $a_1 + a_2 = 1$. It is also assumed that $\widehat{\theta}_{n\in s}$ and $\widehat{\theta}_{n \not\in s}$ are uncorrelated, which is intuitively justified by the fact that the noise contributing to the $n^{th}$ data point is uncorrelated with the noise contributing to any of the previous $n-1$ data points.

The values $a_1$ and $a_2$ in~\eqref{eqn:combination} represent the relative likelihoods of occurrence for the two types of jackknife estimates $\widehat{\theta}_{n\in s}$ and $\widehat{\theta}_{n \not\in s}$, respectively. If we temporarily remove the $n^{th}$ data point from the data set, we see that there are $\binom{n-1}{r}$ possible unique jackknife estimates $\widehat{\theta}_{n \not\in s}$ that can be obtained from $r$ data points. Moreover, it becomes apparent that $\widehat{\theta}_{n \not\in s} = \widehat{\theta}_{n-1}$. Since there are $\binom{n}{r}$ total possible unique jackknife estimates of $\widehat{\theta}_{n}$, the likelihood of reselecting an estimate $\widehat{\theta}_n$ is $\binom{n-1}{r} \binom{n}{r}^{-1} = 1 - r/n$. Hence, one obtains
\begin{equation} \label{eqn:a_soln}
a_1=r/n \text{ and } a_2=1 - r/n.
\end{equation}


By substituting equations~\eqref{eqn:combination} and~\eqref{eqn:a_soln} into~\eqref{eqn:exp_combo}, and observing that $\widehat{\theta}_{n \not\in s} = \widehat{\theta}_{n-1}$, the adaptive jackknife sample mean estimator is defined to be
\begin{equation} \label{eqn:jackknife-adaptive-mean}
\overline{\widehat{\theta}_{n}} = \frac{r}{n} \overline{\widehat{\theta}_{n \in s}} + \left(1-\frac{r}{n}\right) \overline{\widehat{\theta}_{n-1}},
\end{equation}
where
\begin{equation}
\overline{\widehat{\theta}_{n \in s}} = \; \frac{1}{m}\sum_{s\in S_m^+}{\widehat{\theta}_s}, \nonumber 
\end{equation}
and
\begin{equation}
S_m^+ = \left\{ s \in S_m | n+1 \in s = \{i_1,\ldots,i_m\} \right\}. \nonumber
\end{equation} 
Similarly, the jackknife sample variance update is obtained by substituting equations~\eqref{eqn:combination} and~\eqref{eqn:a_soln} into~\eqref{eqn:var_combo}. By again observing that $\widehat{\theta}_{n \not\in s} = \widehat{\theta}_{n-1}$, one gets
\begin{equation} \label{eqn:jackknife-adaptive-var}
\tilde{v}^s_{n} = \left(\frac{r}{n}\right)^2 \tilde{v}^s_{n \in s} + \left(1- \frac{r}{n}\right)^2 \tilde{v}^s_{n-1},
\end{equation}
where
\begin{equation}
\tilde{v}^s_{n \in s} = \frac{r}{(d+1)m}\sum_{s\in S_m^+}{\left(\widehat{\theta}_s - \overline{\widehat{\theta}_{n}} \right) \left(\widehat{\theta}_s - \overline{\widehat{\theta}_{n}}\right)^T}. \nonumber 
\end{equation}

Equation~\eqref{eqn:jackknife-adaptive-mean} inherits the convergence properties of its respective constituent terms $\widehat{\theta}_{n \in s}$ and $\widehat{\theta}_{n \not\in s}$, because each of those constituent terms have identical convergence properties and~\eqref{eqn:jackknife-adaptive-mean} is a convex combination of its constituent terms. The same reasoning about convergence applies to~\eqref{eqn:jackknife-adaptive-var} and its constituent terms $\tilde{v}^s_{n \in s}$ and $\tilde{v}^s_{n \not\in s}$, as well. Furthermore, since we are effectively keeping track of a running average of second-stage $m$ samples, the total number of second-stage samples acquired at measurement number $n=n_0+k$ is $m_n=m_0+km$, where $m_0$ is the number of second-samples used to estimate the first $n_0$ measurements. By choosing $m_0=n_0$, then the condition $m_n=n^\delta$ for some $\delta \geq 1$ is satisfied, and the variance estimate of $v_n$ has the same accuracy as the delete-1 jackknife, but for a fixed number of computations at each increment of $n$.

%

\subsection{Least Squares Parameter Estimator}

The previous sections established general results for the convergence in sample-variance for a parameter estimate without any mention of the parameter estimator. Since we want to make no assumptions about the parameter's prior distribution, we shall choose the well known LSQ estimator. Fortuitously, the LSQ estimator naturally produces a normally distributed parameter estimate~\cite{Shao1992}, which is consistent with the assumed uncertainty terms in the stochastic model~\eqref{eqn:SDEa} and ~\eqref{eqn:SDEb}.

Suppose we have, again, a sequence of $n$ data measurements $D_n=\{Y_1,\ldots,Y_n\}$, where $Y_i=(y_i,t_i)$, as defined earlier. Adopting much of the notation from~\cite{Shao1992}, we consider a general nonlinear model to describe an observed sequence of data
\begin{equation} \label{eqn:nonlin_model}
y_i = H(t_i,\theta)+\sigma e_i, \qquad i=1,\ldots,n,
\end{equation}
where $\theta$ is a vector of unknown constant parameters, $H(t,\theta)$ is a nonlinear function in $\theta$, the $e_i$'s are independent and identically distributed (i.i.d.) unobservable random variables with mean zero and variance one, and $\sigma$ is the unknown error standard deviation. It is also noted that the error terms define the measurement residuals $r_i=(y_i-H(t_i,\theta))=\sigma e_i$. 

A LSQ parameter estimator finds an estimate $\widehat{\theta}_n$ of the parameters that minimizes the mean squared error (MSE) for a model over all available data points
\begin{equation} \label{eqn:LSQ}
\widehat{\theta}_n=\underset{\theta}{\text{argmin}} \frac{1}{n} \sum_{i=1}^n {\left( y_i-H(t_i,\theta) \right)^2},
\end{equation}
which effectively minimizes $\sigma$ in the model \eqref{eqn:nonlin_model}. In relation to the SDE model \eqref{eqn:SDE}, one finds that $H(t,\theta)=h(F(t,x(T)))$ when $\theta=x(T)$ for some fixed point in time $T$.

We remark that the solution to~\eqref{eqn:LSQ} also minimizes the sample variance of the $\widehat{\theta}_n$ estimate's residuals $\text{var}(\widehat{r}_n)$. When using all of the data points, the solution to \eqref{eqn:LSQ} is only one point estimate of the parameters. With only one point estimate of the parameters $\widehat{\theta}_n$, there is no knowledge about how sensitive the parameters are to the data, or equivalently, what the variance estimate is of the parameters (i.e. $\text{var}(\widehat{\theta}_n)$) that produced the given realization of the data. Jackknife variance estimation, such as the JSVE, provides a way of aggregating parameter estimates without making any prior assumptions about the distribution of $\widehat{\theta}$ (i.e., JSVEs are \textit{nonparametric} estimators).


From the given data realization $D_n$, we can implement a delete-d jackknife sampling of $D_n$ to generate a sample distribution of $D$, which directly gives us a sample distribution of $\theta$ by running the LSQ estimator on each jackknife sample of $D_n$. This approach is rigorously studied in~\cite{Shao1992} (and references therein), which specifically describes the asymptotic consistency properties of the LSQ estimator and its jackknife variance estimator in nonlinear models. For the jackknife estimate $\widehat{\theta}_n$ of $\theta_n$, it was found in~\cite{Shao1992} that consistency and asymptotic normality of $\widehat{\theta}_n$ can be established, as well as the consistency of the jackknife variance estimator of the asymptotic covariance matrix of $\widehat{\theta}_n$. The results are summarized here for the delete-1 jackknife, as originally presented in~\cite{Shao1992}, and can easily be extended to the delete-$d$ case using the results of the previous sections.

\begin{itemize}
\item (~\cite{Shao1992} Theorems 1 and 2) For a LSQ estimator $\theta_n$ conditioned on $n$ data points, then $\theta_n \rightarrow \theta$ almost surely (a.s.), and the distribution of a sequence of consistent LSQ estimators $\theta_n$ is asymptotically normally distributed. \\
\item (~\cite{Shao1992} Lemma 3) Let $\widehat{\theta}_{s}$, for $i=\{1,\ldots,n\}$, be the collection of delete-1 jackknife samples of the LSQ estimates of $\theta_n$. Then 
\begin{equation}
\text{max}_{i \leq n}\left\| \widehat{\theta}_{ni} - \theta \right\| \rightarrow 0 \qquad \text{a.s.}
\end{equation}
\item (~\cite{Shao1992} Theorem 4) The jackknife variance estimator is consistent, by proving that $n(\tilde{v}_n - v_n) \rightarrow 0$ a.s.

\end{itemize}

Therefore, a jackknifed sampling of least squares estimates allows us to estimate a prior distribution of parameters for a nonlinear model without having to implement MCMC methods. An added benefit of the LSQ jackknife sampling procedure is that the estimated parameter distribution will asymptotically be normally distributed. Ensuring that the distributions are normal is essential to the performance of the EnKF, since the EnKF only uses the first two moments of the ensemble distribution. Furthermore, the adaptive scheme in the previous section provides a computationally efficient way of assimilating new data into the statistical model.

\section{Posterior estimation via ensemble filtering} \label{sec:QandR}

In previous sections, we saw how to use ensemble filtering to construct posterior estimates of a distribution's mean and covariance without having to implement MCMC methods. However, the ensemble filtering requires knowledge of the process noise, measurement noise, and the mean and covariance of the prior distribution. When these prior quantities are known, ensemble filtering can be implemented to further reduce the computational cost of assimilating new data. Without prior knowledge of model parameters or model noise distributions, we propose that one can implement jackknife estimation methods to initialize the stochastic model such that ensemble filtering can take over the posterior parameter and state estimation process once it produces posterior estimates that agree with that of the adaptive jackknife method.

\subsection{Estimating R from Cross-Validation}

When implementing the jackknife LSQ estimator, the sampling distribution for $\theta$ produces an output distribution for $y$. However, the measurements are subject to uncertainty, as accounted for in~\eqref{eqn:SDE}, and this uncertainty can be measured as being attributed to the additional \textit{out-of-sample} error. Cross-validation (CV) is a statistical learning technique typically used to evaluate a model by describing its out-of-sample statistics. Typical CV methods involve training a model on a subset $S_m$ of the available data, and then validating (testing) the model on the complement of $S_m$, which we denote as $S_m^c$.

The delete-$d$ jackknife variance estimator already removes $d$ data points from the available data before each step of the parameter estimation, which naturally allows us to use those $d$ data points to acquire out-of-sample residual statistics that are indicative of the errors we would see for a future measurement. Furthermore, we can use the delete-$d$ jackknife methodology to obtain jackknife estimates of the residual statistics, except the validation set uses a delete-$r$ jackknife estimate.

For a given jackknife parameter estimate $\widehat{\theta}_s$ such that $s \in S_m$, a residual $\widehat{r}_j$ is defined for some $j \in S_m^c$ as
\begin{equation}
\widehat{r}_j = y_j - H(t_j,\widehat{\theta}_s),
\end{equation}
and the residual statistics defined for a set of $\mu$ indices $\left\{j_1, \ldots, j_{\mu} \right\} \in S_m^c$, for which $\mu \leq d$, are
\begin{align*}
\overline{\widehat{r}_s} &= \frac{1}{\mu}\sum_{j \in S_m^c}{\widehat{r}_j}, \\
\widehat{\sigma}^2_s &= MSE(\widehat{\theta}_{s}) = \frac{d}{r \mu}\sum_{j \in S_m^c}{  \widehat{r}_j \widehat{r}_j^T},
\end{align*}
where $\widehat{\sigma}^2_s$ estimates the out-of-sample variance of $\widehat{\theta}_s$.

For each $\widehat{\theta}_s$ estimate, there exists a corresponding jackknife sample distribution of out-of-sample residual values $\widehat{r}_j$. The jackknife mean of $\widehat{r}$ is the measurement bias, and the jackknife variance estimate $\widehat{\sigma}^2_s$ captures the uncertainty attributed to both $\widehat{\theta}_n$ and the measurement noise $\sqrt{R}N(0,1)$. Since we obtain $m$ estimates of $\widehat{\theta}_s$, we also obtain $m$ sample distributions of the out-of-sample residuals, and the expected residual distribution is described by the arithmetic mean of the $m$ residual distributions (i.e., each residual distribution has equal probability of being the correct residual distribution). The expected jackknife residual statistics are
\begin{align}
\overline{\widehat{r}_n} &= \frac{1}{m}\sum_{s \in S_m^c}{\widehat{r}_s}, \label{eqn:resid_mean} \\
\widehat{\sigma}^2_n &= \frac{1}{m^2}\sum_{s \in S_m^c}{\widehat{\sigma}^2_s}. \label{eqn:resid_var}
\end{align}
The adaptive rule outlined in Section \ref{sec:jackknife} can also be applied to obtain
\begin{align}
\overline{\widehat{r}_n} &= \frac{r}{n}\widehat{r}_{n \in s^c} + \left(1-\frac{r}{n}\right)\overline{\widehat{r}_{n-1}} \label{eqn:adaptive_bias} \\
\widehat{\sigma}^2_n &= \left(\frac{r}{n}\right)^2 \widehat{\sigma}^2_{n \in s^c} + \left(1-\frac{r}{n}\right)^2 \widehat{\sigma}^2_{n-1}, \label{eqn:adaptive_sigma}
\end{align}
where $\widehat{r}_{n \in s^c}$ and $\widehat{\sigma}^2_{n \in s^c}$ are defined by~\eqref{eqn:resid_mean} and~\eqref{eqn:resid_var} with $n \in S_m^c$.

By taking $P_y = \sigma^2_n$, and 
\begin{equation}
\widehat{P}_y = \frac{r}{dm}\sum_{s\in S_m}{\left( H(t_s,\widehat{\theta}_s)- \frac{1}{m}\sum_{s \in S_m}{H(t_s,\widehat{\theta}_s}) \right)\left([\cdots]\right)^T}, \label{eqn:LSQ_meas_var}
\end{equation}
then one can solve for $R$ from \eqref{eqn:Py_ensemble} to get
\begin{equation} \label{eqn:R_soln}
R = \widehat{\sigma}^2_n - \widehat{P}_y.
\end{equation}
Because the LSQ estimator finds a deterministic realization of each $\widehat{\theta}_s$ assuming no stochastic forcing, it is noted that when using the definitions \eqref{eqn:resid_var} and \eqref{eqn:LSQ_meas_var}, the $\widehat{Q}_y$ term of \eqref{eqn:R_soln} is identically equal to the zero matrix. It is also noted that by combining the results of~\cite{Shao1989b} and~\cite{Shao1992}, both $\widehat{\sigma}^2_n$ and $\widehat{P}_y$ are each aymptotically consistent, and thus $R$ is asymptotically consistent as well.

One can also account for the measurement bias in~\eqref{eqn:(10.97a)} to correct the expected output signal
\begin{equation} \label{eqn:(10.97a)-unbiased}
\overline{\widehat{x}^+} = \overline{\widehat{x}^-} + K\left(y-\overline{y} - \overline{\widehat{r}_n} \right).
\end{equation}

\subsection{Estimating Q from the ensemble filter}

When comparing the jackknife LSQ estimator model \eqref{eqn:jackknife-1a} to the SDE model \eqref{eqn:SDE}, the parameter vector $\theta$ of the LSQ estimator is usually comprised of the SDE state values $x(t)$ at some time $t_k$: 
\begin{equation}
\theta_k = \begin{pmatrix}x(t_k) \\ x_p \end{pmatrix},
\end{equation} 
where the SDE state values $x(t)$ are augmented by the SDE model parameters $x_p$ having zero deterministic dynamics (i.e., $dx_p=Q_pdw$). It follows from \eqref{eqn:SDE_soln} that
\begin{equation} \label{eqn:theta_to_x}
\begin{bmatrix} x(t_{k+1}) \\ x_p \end{bmatrix} = \begin{bmatrix} F(t_{k+1},\theta_k) \\ 0 \end{bmatrix} + \sqrt{Q_k \Delta t_k}N(0,1).
\end{equation}
Here,
\begin{equation*}
Q_k = \begin{bmatrix} Q_t & Q_{tp} \\ Q_{tp} & Q_p \end{bmatrix},
\end{equation*}
where $Q_t$ is the $Q$ from \eqref{eqn:SDE_soln}, $Q_p$ is the auto-covariance of uncertainty in the parameters, and $Q_{tp}$ represents the cross-covariance between uncertainty in the states and parameters. Together, $Q_k$ defines the process uncertainty of the augmented stochastic model \eqref{eqn:theta_to_x}.

By treating each jackknife estimate $\widehat{\theta}_s$ as an ensemble estimate $\widehat{\theta}_i$, we have $N$ ensemble estimates at time $t_k$:
\begin{align}
\widehat{x}^{(i)} &= F(t_k,\widehat{\theta}_{ki}), \\
\widehat{P}_x &= \frac{1}{N}\sum{(\widehat{x}^{(i)} - \overline{\widehat{x}^{(i)}}) ([\cdots])^T}.
\end{align}

Essentially, the jackknife samples represent an ensemble of state estimates via the transformation of \eqref{eqn:theta_to_x}. In terms of the ensemble filtering framework, the posterior state covariance matrix $P_x^+$ for state values $x(t_k)=\theta_k$ can be estimated from the jackknife variance estimate $v_{k}$, and the prior state covariance matrix $P_x^-$ for the state values $x(t_k)=\theta_{k-1}$ can be estimated by evolving ensemble members backward in time to $t_k$ (similar to the prediction step in UKF) and calculating the ensemble variance at that time step, say $\widehat{P}_x^-$. The justification here is that both $P_x^+$ and $v_{n}$ are representations of the state covariance matrix after assimilating new data. By substituting $v_{n} = \widehat{P}_x^+$ into \eqref{eqn:ensemble_var} and taking $\widehat{\sigma}^2_{n} = \widehat{P}_y + R$, one can explicitly solve for $Q$:
\begin{equation} 
Q = \frac{1}{\Delta t_k} \left( v_n -\widehat{P}_x^- +\widehat{P}_{xy}(\widehat{\sigma}^2_n-\widehat{Q}_y)^{-1}\widehat{P}_{xy}^T \right). \label{eqn:Q_solve}
\end{equation}

\subsection{Discussion}

To simplify the computation of \eqref{eqn:Q_solve}, the $\widehat{Q}_y$ can be omitted from \eqref{eqn:Q_solve}, which will yield a pessimistic (i.e., greater in norm) solution for $Q$ since $\widehat{Q}_y$ is positive semi-definite and contributes positively to an inverted term. For many applications, including robust control, this is an acceptable approximation.

For highly nonlinear systems, the LSQ procedure may possibly find a region of minima that are located significantly further away from the dominant mode. These types of secondary modes can quickly emerge and cause the $\widehat{P}_x^-$ to be large enough to make \eqref{eqn:Q_solve} negative semi-definite. One solution to this problem would be to implement a Gaussian mixture model (GMM) on the ensemble of realizations and run the adaptive Kalman filter on the constituent normal distributions of ensemble members. In cases where this approach is too computationally costly, the $\widehat{P}_x^-$ term can be omitted from \eqref{eqn:Q_solve} to, again, yield an even more pessimistic solution for $Q$.
\section{Example application} \label{sec:example}


To demonstrate the performance of the adaptive jackknife estimator, we shall consider a simple logistic model with discrete measurements and additive noise:
\begin{align}
\begin{pmatrix} dx \\ d\beta \\ dN \end{pmatrix} &= \begin{pmatrix} \beta x \left(1 - \frac{x}{N}\right) \\ 0 \\ 0 \end{pmatrix}dt + \sqrt{Q}dw \label{eqn:example_ODE}\\
y_k &= x(t_k) + \sqrt{R}N(0,1), \label{eqn:example_meas}
\end{align}
where $\beta \in \mathbf{R^+}$ is the growth parameter, $N \in \mathbf{R^+}$ is the upper bound of $x$, and $dq$ and $\sqrt{R}N(0,1)$ are the noise processes described in Section~\ref{sec:statistical_kalman}. The logistic model defined by~\eqref{eqn:example_ODE} and~\eqref{eqn:example_meas} is a common model used to describe the adoption of a behavior or new technology~\cite{Hethcote2000}, and is known to have well known convergence properties when using a jackknife sampling LSQ variance estimator~\cite{Shao1992}. It is also noted that the integral of the deterministic part of~\eqref{eqn:example_ODE} (i.e., $dx=\beta x \left(1 - \frac{x}{N}\right)dt$) has the solution:
\begin{equation}
x(t)=\frac{N x(0) \exp{\beta t}}{N+x(0)\left(\exp{\beta t}-1\right)}. \label{eqn:example_soln}
\end{equation}

\begin{figure}[t]
  \centering
    \includegraphics[width=0.48\textwidth]{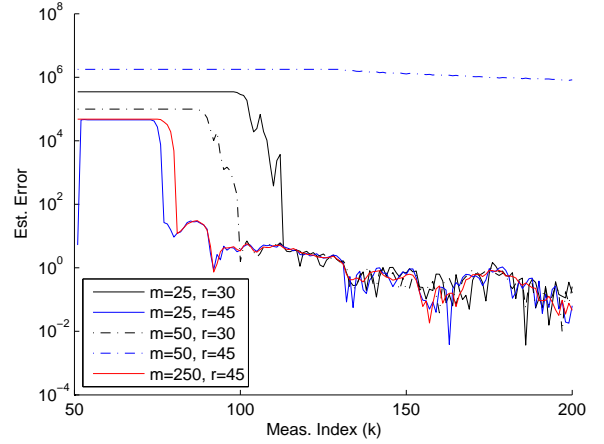} 
  \caption{Adaptive jackknife estimation performance evaluation for a logistic model, with different jackknife parameter values. In all test cases, $n=50$ and $\mu=n-r$.}\label{fig:trackingA}
\end{figure}

We simulated a sequence of 200 measurements $y_k$, at times uniformly distributed on the interval $t=[0,80]$, with initial values $\left(x(0),\beta,N\right)=\left( 1,   0.225,  500\right)$, and noise covariance matrices $Q=\text{diag}(15,0.001,10)$ and $R=1$. Figure~\ref{fig:trackingA} shows the error, in Euclidean norm, between the state estimate of the adaptive jackknife filter and the value of~\eqref{eqn:example_soln} at time $t_k$. For a fixed \textit{burn-in} period of 50 measurements, we find that the estimate of the augmented state vector converges with a greater number of included measurements $r$, and fewer jackknife samples $m$. With a greater value of $r$, the adaptive jackknife filter is able to use as many measurements as possible during the burn-in initialization phase, which $(i)$ causes a reduction in the jackknife variance estimate according to~\eqref{eqn:jackknife-adaptive-var}, and $(ii)$ results in an initial estimate closer to the true value by causing the value $r/n$ to be large. Using fewer jackknife samples (i.e., $m=25$ vs $m=50$) seems to also counter-intuitively produce a fast convergence result in this example, because few jackknife samples are needed to accurately represent the uncertainty distributions in the model. Choosing $m=50$ causes an over-sampling of outliers, and it is not until we have $m=250$ that the true distribution emerges.


\section{Conclusion and Future Work} \label{sec:conclusion}

We have shown how one can implement the techniques of jackknife variance estimators as they apply to least squares estimators to construct an adaptive, nonparametric, and computationally efficient statistical nonlinear filter. 


One issue that we left as an assumption is that for each jackknife estimate, there exists a solution to the LSQ problem. In fact, this is not a far-fetched assumption to make because bootstrap methods (similar to jackknife sampling) have been shown to efficiently search for the solution to the general LSQ problem \cite{James2013}. Lastly, we also remark that jackknife sampling LSQ problem is easily broken down to a parallel computation problem, since the LSQ solution for each jackknife sample of the data can be solved independently of each other jackknife sample. Therefore, there is room for future work on this subject to increase computational efficiency, both with respect to improving LSQ estimation and parallelizing each step of the adaptive algorithm.

\addtolength{\textheight}{-3cm}   
\section{ACKNOWLEDGMENTS}

This work was supported by the Institute for Collaborative Biotechnologies through grant W911NF-09-0001 from the U.S. Army Research Office and by the Army Research Laboratory under cooperative agreement W911NF-09-2-0053 (NS-CTA). The content of the information does not necessarily reflect the position or the policy of the Government, and no official endorsement should be inferred. The U.S. Government is authorized to reproduce and distribute reprints for Government purposes notwithstanding any copyright notice herein.



%
%
%
%

\bibliographystyle{ieeetr}

\end{document}